\documentclass{article}
\usepackage{arxiv}

\usepackage[utf8]{inputenc} 
\usepackage[T1]{fontenc}    
\usepackage{hyperref}       
\usepackage{url}            
\usepackage{booktabs}       
\usepackage{amsfonts}       
\usepackage{nicefrac}       
\usepackage{microtype}      
\usepackage{lipsum}
\usepackage{graphicx}
\graphicspath{ {./images/} }
\usepackage{natbib}

\usepackage{xcolor}
\hypersetup{
    colorlinks=true,
    citecolor=teal,
    linkcolor=teal,
    filecolor=teal,
    urlcolor=teal,
}
\usepackage[capitalize, nameinlink]{cleveref}

\title{Human Cognitive Biases in\\ Explanation-based Interaction:\\ The Case of Within and Between Session Order Effect}

\author{
 Dario Pesenti \\
  CIMeC, University of Trento\\
  \texttt{dario.pesenti@unitn.it} \\
   \And
 Alessandro Bogani \\
  CIMeC, University of Trento\\
  \texttt{alessandro.bogani@unitn.it} \\
  \AND
  Katya Tentori \\
  CIMeC, University of Trento\\
  \texttt{katya.tentori@unitn.it} \\
  \And
  Stefano Teso \\
  CIMeC \& DISI, University of Trento\\
  \texttt{stefano.teso@unitn.it} \\
}

\begin{document}
\maketitle

\begin{abstract}
    Explanatory Interactive Learning (XIL) is a powerful interactive learning framework designed to enable users to customize and correct AI models by interacting with their explanations.
    In a nutshell, XIL algorithms select a number of items on which an AI model made a decision (\textit{e.g.} images and their tags) and present them to users, together with corresponding explanations (\textit{e.g.} image regions that drive the model’s decision). Then, users supply corrective feedback for the explanations, which the algorithm uses to improve the model.
    Despite showing promise in debugging tasks, recent studies have raised concerns that explanatory interaction may trigger \textit{order effects}, a well-known cognitive bias in which the sequence of presented items influences users’ trust and, critically, the quality of their feedback.
    We argue that these studies are not entirely conclusive, as the experimental designs and tasks employed differ substantially from common XIL use cases, complicating interpretation.
    To clarify the interplay between order effects and explanatory interaction, we ran two larger-scale user studies ($n = 713$ total) designed to mimic common XIL tasks.  Specifically, we assessed order effects both \textit{within} and \textit{between} debugging sessions by manipulating the order in which correct and wrong explanations are presented to participants.
    Order effects had a limited, though significant, impact on users' agreement with the model (\textit{i.e.}, a behavioral measure of their trust), and only when examined within debugging sessions, not between them. The quality of users' feedback was generally satisfactory, with order effects exerting only a small and inconsistent influence both within and between sessions.
    Overall, our findings suggest that order effects do not pose a significant issue for the successful employment of XIL approaches. More broadly, our work contributes to the ongoing efforts for understanding human factors in AI.\footnote{{Our study has received approval from the Ethics board of our university.}}
\end{abstract}


\section{Introduction}


Explainable AI (XAI) tools help stakeholders inspect, understand, and evaluate the behavior of -- otherwise opaque -- AI models by generating \textit{explanations} of their decisions \citep{molnar2020interpretable, schwalbe2024comprehensive}.
These can often reveal defects in the model's reasoning that would be difficult to detect through standard evaluation alone, such as sub-optimal feature usage \citep{kulesza2015principles} and reliance on confounded features \citep{geirhos2020shortcut, lapuschkin2019unmasking, ye2024spurious}.
\textit{Explanatory interactive learning} (XIL) is a powerful framework that builds on this observation to improve user control \citep{kulesza2015principles, teso2019explanatory}.
XIL algorithms repeatedly select a number of items on which an AI model made a decision (\textit{e.g.} images and their tags) and present them to users together with corresponding local explanations (\textit{e.g.} highlighting what regions of the image drove the model's decision). Then, they ask users to evaluate whether the provided explanations are satisfactory and, if not, to supply corrective feedback, which the algorithms use to improve the AI model.
By steering directly the model's explanations, XIL enables users to quickly debug and customize AI models \citep{lertvittayakumjorn2021explanation, teso2023leveraging}; see \cref{sec:related-work} for a broader overview.

Recent studies \citep{nourani2021anchoring} have shown that the order in which AI predictions are presented to users may affect how users interact with them. 
\textit{Order effects} refer to a class of cognitive biases \citep{HOGARTH19921} in which human judgments are systematically influenced by the sequence in which information is presented. 
More specifically, a \textit{primacy} effect occurs when earlier information is given disproportionately more weight than more recent information, whereas a \textit{recency} effect arises when the opposite happens, with more recent information exerting greater influence.
In the context of intelligent systems, \citet{nourani2021anchoring} have shown that, in an non-XIL interactive setting, users witnessing correct AI behaviour early tend to build a more accurate mental model of the AI but also to \textit{over-rely on its suggestions}
and to \textit{make more mistakes} when prompted to decide themselves, while those that encounter errors early tend to underestimate the AI's competencies.  This suggests that, in XIL, \textit{order effects may compromise the quality of human feedback and thus its reliability in applications}.

\begin{figure*}
    \centering
    \includegraphics[width=0.88\textwidth]{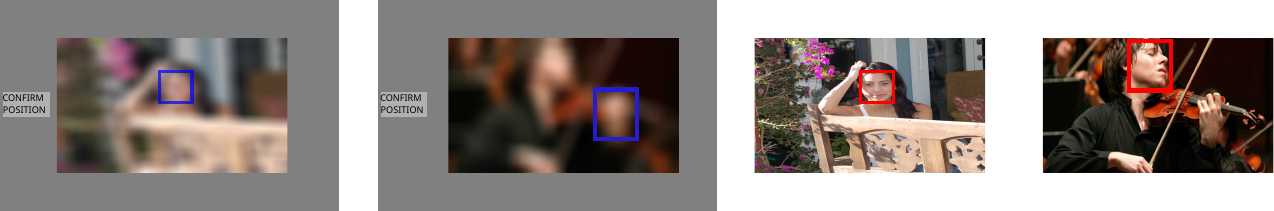}
    \caption{\textbf{Interface of the user studies} (left): participants were asked to evaluate the model's \textit{explanation} (represented by a bounding box, in blue) of a fictional image classifier. Ideally, this should entirely enclose the face of the person pictured in the image.  Participants were instructed to press the ``Confirm'' button if they deemed the box enclosed the face, or to move it onto the face otherwise, all within a 6 seconds time limit. Right: original images and ground-truth bounding boxes (in red), for reference.}
    \label{fig:page-2}
\end{figure*}

We contend that the question is not yet settled, as \citeauthor{nourani2021anchoring}'s study differs substantially from typical XIL use cases (cf. \cref{sec:preliminaries} for a discussion). Their findings also partially conflict with those of \citet{honeycutt2020soliciting}, who, in a more interactive design, found a detrimental impact of letting users interact with the model on their perception of it, but no significant order effects.

To understand the interplay between explanatory interaction and order effects, we carry out two controlled, larger scale user studies ($n=713$ total) simulating a XIL task.
During the experiments, participants had to interact with explanations from an image classifier that was fictitious, unbeknownst to them. More specifically, participants were instructed to either accept the model's explanation, or to correct it (cf. \cref{fig:page-2}).

We considered two settings representative of actual XIL usage:
in the \textbf{within-session} setting, we focused on order effects occurring within a single debugging session, before the AI model was updated;
whereas in the \textbf{between-session} setup, participants carried out two consecutive debugging sessions, believing that the AI model had been updated in between.
Our results
indicate that the order of presentation of AI outputs exerted only small effects on the quality of users' feedback, which presented overall high levels, in both within- and between-session settings, as well as on their implicit agreement with the model. Only a small order effect was found within-session. No effect was found on users' perceptions of accuracy and trustworthiness of the model.
In summary, \textit{order effects do not appear to pose a significant issue for the successful employment of XIL approaches}.  Our work contributes to ongoing efforts for understanding human factors in AI and explanation-based interaction.

\section{Preliminaries}
\label{sec:preliminaries}

%
%
%
%
%

\paragraph{Explanations and interaction.}  \textit{Explanatory interactive learning} (XIL) operationalizes the observation that if a (sufficiently expert) user understands how a model works, they can -- and, typically, proactively want to \citep{kulesza2015principles} -- supply corrective feedback useful for improving the model itself \citep{teso2019explanatory, schramowski2020making}.
For instance, in a medical diagnosis task using X-ray scans, machine learning classifiers can achieve high accuracy by exploiting confounding factors, \textit{e.g.} background cues that correlate with the decision but are not causally related.  This compromises out-of-sample performance while being invisible to standard metrics like accuracy.  By highlighting what features the model relies on, local explanations such as saliency maps \citep{miller2019explanation} help identify such issues \citep{geirhos2020shortcut, lapuschkin2019unmasking} and enable users to formulate corrective feedback (\textit{e.g.} ``don't use this part of the scan'') \citep{ross2017right}.

XIL algorithms loop through two steps.
During the \textbf{debugging session}, the machine iteratively selects a number of items (\textit{e.g.} images) from a pool of options and computes predictions (tags) and explanations (saliency maps) for them.  The items and the corresponding predictions and explanations are then presented to a user, who, for each of them, can indicate what features the machine is using improperly.
In the \textbf{update step}, the collected feedback is used to update the model.
XIL approaches were shown to, \textit{e.g.} help laypeople to quickly tailor spam filters to their needs \citep{kulesza2015principles} and domain experts to rectify confounding in scientific studies \citep{schramowski2020making}.
%
%
However, it has been mostly neglected whether human biases -- such as order effects, which are naturally triggered by the XIL loop -- can impair the adoption of XIL algorithms.


\paragraph{Order effects and XIL.} 
Research in psychology has shown that order effects are pervasive and impactful, influencing item recall \citep{FURNHAM201135, baddeley1993recency}, belief updating \citep{HOGARTH19921}, response accuracy \citep{eisenberg1988order}, and even preferences in political elections \citep{matsusaka2016ballot}.
They have received increasing attention also in the XAI literature \citep{nourani2021anchoring, nourani2020role, nourani2022importance}. Indeed, both within- and between-session order effects can influence the quality of debuggers’ performance, and each requires ad hoc solutions. For example, the former can be addressed by randomizing the order of item presentation, whereas the latter would necessitate recruiting different groups of debuggers across sessions. It is therefore essential to ascertain the extent to which such effects affect performance; nevertheless, only a limited number of studies have investigated them within interactive learning contexts.
%
%
\citet{honeycutt2020soliciting} showed that allowing the users to correct and interact with a model reduced their trust in it and perception of its accuracy overall, but they found no order effects on trust over multiple debugging sessions. However, they examined only between-session order effects, and focused exclusively on self-report measures, which are less reliable than behavioral indices of accuracy \citep{warren2024categorical} and are often poor predictors of individuals' actual behaviour \citep{sheeran2016intention}.
%
In another work, \citet{nourani2021anchoring} reported an increased mistrust in the model if they were exposed to more incorrect decisions first; conversely, they found an automation effect (\textit{i.e.} over-reliance on model predictions \citep{cummings2012, rastogi2022deciding}) when the users were presented with more correct AI decisions first, which was also reflected in a worse task performance by the participants. In this case, however, they studied only within-session order effects where participants, unlike typical XIL use cases, self-selected the stimuli to inspect and could not correct the model’s outputs.



%

\section{User Studies}
\label{sec:methods}

Our experiments aim to provide a more reliable exploration of the potential negative impact of order effects on XIL algorithms.  Specifically, we investigate the influence of order effects both \textit{within-session} (Experiment 1) and \textit{between-sessions} (Experiment 2), evaluating not only users’ perceptions, but also \textit{their actual behaviour} in a more ecological debugging task.
Next, we introduce the task, variables and data processing common to both experiments.


\paragraph{Task.}
Based on the work of \citep{honeycutt2020soliciting}, we implemented a debugging task of a binary classifier trained to determine the presence of human faces in noisy images. In this task, that comprised either a single (Experiment 1) or two debugging sessions (Experiment 2), participants were shown a series of blurred images containing a face, each accompanied by a bounding box indicating the face location according to the model (\textit{i.e.} the explanation for what part of the image led the model to categorize the image as presenting a face).\footnote{We chose face recognition as a use case because it is accessible, requiring little domain knowledge, yet sufficiently complex to vary stimulus difficulty and avoid floor or ceiling effects.}
The box was placed either correctly or incorrectly (\textit{i.e.} perfectly around the human face or not; see below for further details) and participants were asked to provide feedback about it: if they agreed with it, they had to click a button to confirm its position; conversely, if they deemed the box misplaced, they had to move the bounding box to the position they considered correct with the mouse (the interface is shown in \cref{fig:page-2}.) Importantly, participants' could change only the position of the box, but not its dimensions. For each image, participants had $6$ seconds to provide their feedback: if no response was given within this time frame, the trial was recorded as a ``missed'' response. 

\paragraph{Images and correct bounding boxes.}
All images were sourced from the Open Images Dataset V7 \citep{OpenImages} a well-known repository of natural images.
We selected images containing a single unoccluded, front facing human face, and recorded them along with the face's correct ``ground-truth'' bounding boxes provided by OpenImages.
The selected images were applied a substantial amount of Gaussian blur ($85 \times 85$ kernel, $\sigma = 40$). This was done to make the task less trivial compared to the one employed in \citet{honeycutt2020soliciting}, as a task in which the model's outputs are clearly correct or incorrect might reduce the chances of observing possible order effects (as partially supported by our findings, see \cref{sec:experiment-within}).

\begin{figure}
    \centering
    \includegraphics[width=0.65\linewidth,keepaspectratio]{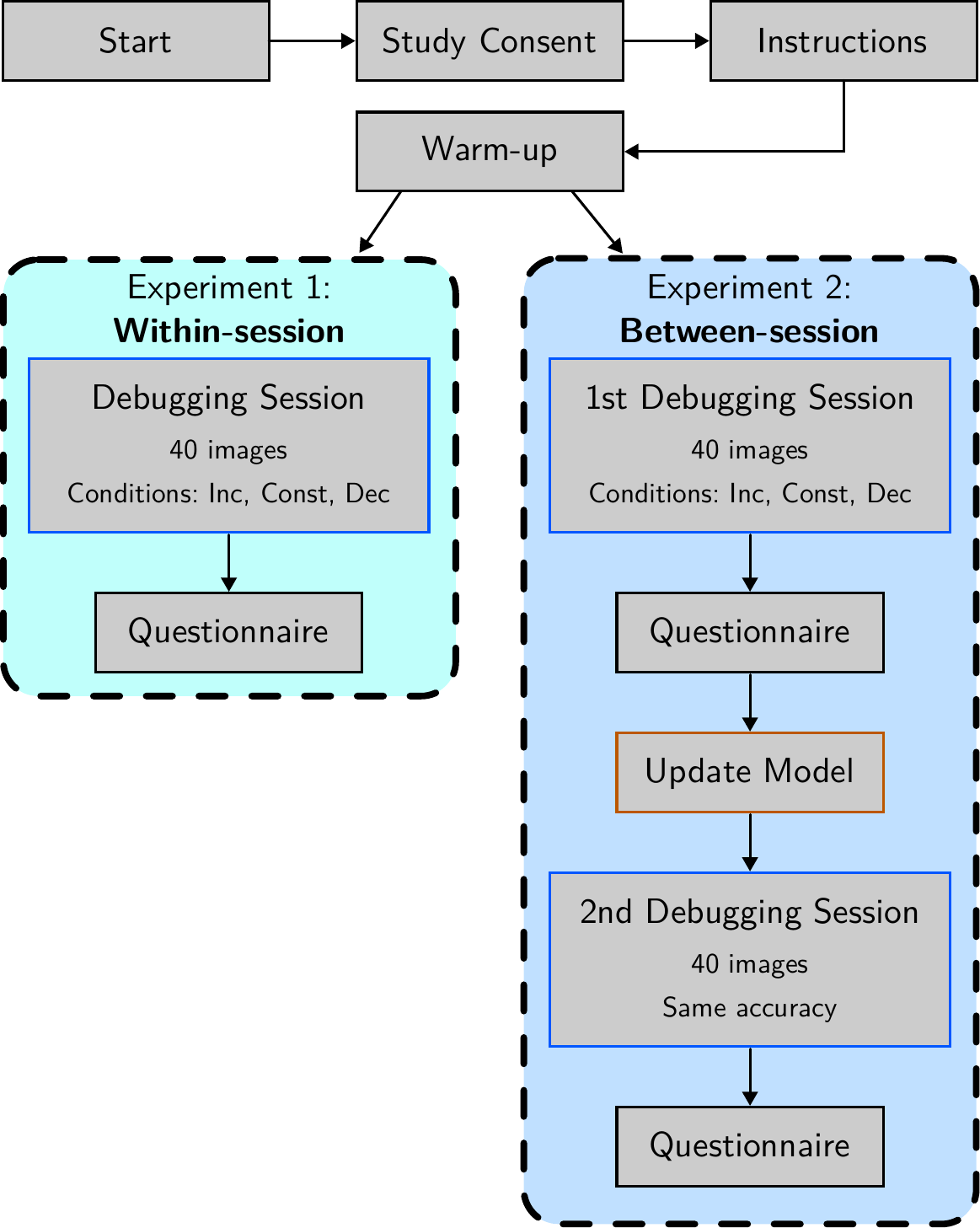}
    \caption{Schematic illustration of our user studies.}
    \label{fig:flow-chart}
\end{figure}

\paragraph{Variables and Analyses.} In both experiments, we manipulated the following three independent variables.
\begin{enumerate}
    \item \textit{Order}: It refers to how the accuracy of the model (represented by the proportion of images featuring a correct box) evolved throughout the debugging task (\textit{i.e.} between the first and second half of the single session in Experiment 1 and between the two sessions in Experiment 2; see \cref{tab:exp1-conditions}). It presented 3 levels (between subjects): increasing (\textit{i.e.} accuracy improved throughout the task), constant (\textit{i.e.} accuracy remained constant), and decreasing (\textit{i.e.} accuracy worsened), abbreviated as \textit{Inc}, \textit{Const}, and \textit{Dec}, respectively.

    \item \textit{Placement}: It refers to the accuracy of the model’s placement of the bounding box, defined as the percentage of overlap between model-placed boxes and the ground truth.  It presented 2 levels (within-subject): \textit{correct} vs. \textit{incorrect}. Correct placements consist in bounding boxes coinciding perfectly with the ground truth. Among the incorrect placements, we distinguish between two categories: ‘partially wrong’ (with a 25\% overlap between the model’s placement and the ground truth) and ‘wrong’ (with no overlap at all). We introduced this distinction to represent the variety of errors that a model might make, but due to the low number of stimuli belonging to each of this two sub-classes (see \cref{tab:exp1-conditions}) they were considered as a single ``incorrect'' class in the statistical analyses (but see Appendix B for plots of the results split by the three placement levels).\footnote{False positives, where the model detects a non-existent face, were excluded, as XIL methods chiefly target models that make correct predictions for the wrong reasons \citep{teso2023leveraging}.}

    \item \textit{Difficulty}: It refers to how challenging it was to locate the human face in the image, and it presented 2 levels (within-subjects): \textit{easy} vs. \textit{difficult}. This variable was included to assess whether order effects, if present, influenced responses across all stimuli or specifically in those where the correct answer was more ambiguous. Image difficulty was determined through pilot studies in which candidate images (together with their respective bounding boxes) were presented to participants who had to confirm or correct the position of the box. Easy and difficult images were selected, respectively, among the ones presenting the highest and lowest levels of accuracy (calculated as the percentage of overlap between participant-placed boxes and the ground truth). 
\end{enumerate}

In both experiments, participants across all conditions viewed the same set of images. Thus, any differences between the Inc and Dec groups could be attributed solely to presentation order, with each group effectively serving as a control for the other. The inclusion of the Const group, in which no order effects were present, provided a baseline condition that enabled an even more precise interpretation of potential differences between the other two groups.
%



%
We then measured three dependent variables:
\begin{enumerate}

    \item \textit{Accuracy of participants' feedback}, quantified as the overlap between the area corresponding to the ground truth and that of
participants’ placement, divided by the former;

    \item \textit{Participants' agreement with the model}, quantified as the overlap between the areas of the model’s box and the participants’ box, divided by the former;

    \item \textit{Participants' perception of the model’s accuracy and their trust in it}, quantified through a questionnaire comprising four 7-point Likert scale items adapted from \citet{honeycutt2020soliciting} and \citet{hoffman2019metrics}.
        
\end{enumerate}
The exact wording of the task instructions and the questionnaire items is reported in \cref{sec:questionnaire-items}.

The accuracy and agreement of the participants with the model were analyzed using mixed linear models that present the order condition, the placement of the model, the difficulty of the image, and their interactions as fixed effects. Random intercepts were included for both participants and images. Significant fixed effects involving more than one contrast between groups were further investigated through Bonferroni-corrected post-hoc comparisons. Participants' answers to the four questionnaire items were averaged to compute an index of perceived model accuracy and trustworthiness and analyzed by means of a Kruskal-Wallis rank sum test to investigate potential differences between the three order conditions.

\paragraph{Participants and data exclusions.} An a priori power analysis, conducted by means of a simulation approach implemented in R \citep{green2016simr, kumle2021estimating}, indicated that, for both experiments, a total sample of at least 330 participants evaluating 40 images each would provide 82\% power to detect a small-to-medium effect of the interaction among the three independent variables, as well as their main effects. Participants were recruited on Prolific\footnote{www.prolific.com} among those with a Prolific approval rate of at least 98\%, and compensated in accordance with the hourly payment suggested by Prolific (£1 for Experiment 1 and £1.30 for Experiment 2, plus a possible bonus payment of £10;  see \cref{sec:experiment-within}).

Before running the analyses, we excluded all observations in which participants did not confirm nor changed the position of the box, as well as those in which the last registered input was made at $5.95$ sec or later from the onset of the image out of the total $6$ seconds available, as these likely reflected cases in which participants had not reached a conclusive answer by the end of the trial.\footnote{Still, running the analyses without excluding this second class of stimuli did not change the results. }

\subsection{Experiment 1: Within-session Order Effect}
\label{sec:experiment-within}

\paragraph{Aim.}
In Experiment 1, we investigated \textit{whether the distribution of the model’s errors within a single debugging session influenced how participants interacted with and perceived the model}. More specifically, we examined whether a prevalence of the model’s errors either at the beginning or at the end of the session, while keeping the model’s overall accuracy exactly the same, affected participants’ performance in the debugging task and their trust in the model.

\begin{table*}[!t]
    \caption{\textbf{Number of images presented in Experiment 1 }(within-session, top) \textbf{and 2}  (between-sessions, bottom), split by Condition (Increasing, Constant, and Decreasing model performance over time), by Image Difficulty (Easy vs. Hard), and by Correctness of the model’s explanation (Correct, Partially Wrong, and Wrong).}
    \label{tab:exp1-conditions}
    \centering
    \small
    \resizebox{\dimexpr0.885\textwidth}{!}{%
    \begin{tabular}{ccccccccccccc}
    \toprule
    \textbf{Within-session}
        & \multicolumn{6}{c}{\textbf{First half of the session}}
        & \multicolumn{6}{c}{\textbf{Second half of the session}}
    \\
        & \multicolumn{3}{c}{\textbf{Easy}}
        & \multicolumn{3}{c}{\textbf{Hard}}
        & \multicolumn{3}{c}{\textbf{Easy}}
        & \multicolumn{3}{c}{\textbf{Hard}}
    \\
    \cmidrule(lr){2-4}
    \cmidrule(lr){5-7}
    \cmidrule(lr){8-10}
    \cmidrule(lr){11-13}
    \textbf{Condition}
        & Corr.
        & P. Corr.
        & Wrong
        & Corr.
        & P. Corr.
        & Wrong
        & Corr.
        & P. Corr.
        & Wrong
        & Corr.
        & P. Corr.
        & Wrong
    \\
    \cmidrule(lr){1-1}
    \cmidrule(lr){2-7}
    \cmidrule(lr){8-13}
    Increasing
        & 4 & 3 & 3 & 4 & 3 & 3 & 8 & 1 & 1 & 8 & 1 & 1
    \\
    Constant
        & 6 & 2 & 2 & 6 & 2 & 2 & 6 & 2 & 2 & 6 & 2 & 2
    \\
    Decreasing
        & 8 & 1 & 1 & 8 & 1 & 1 & 4 & 3 & 3 & 4 & 3 & 3
    \\
    \bottomrule
\end{tabular}
    }
    \resizebox{\dimexpr0.885\textwidth}{!}{%
    \begin{tabular}{ccccccccccccc}
    \toprule
    \textbf{Between-session}
        & \multicolumn{6}{c}{\textbf{First session}}
        & \multicolumn{6}{c}{\textbf{Second session}}
    \\
        & \multicolumn{3}{c}{\textbf{Easy}}
        & \multicolumn{3}{c}{\textbf{Hard}}
        & \multicolumn{3}{c}{\textbf{Easy}}
        & \multicolumn{3}{c}{\textbf{Hard}}
    \\
    \cmidrule(lr){2-4}
    \cmidrule(lr){5-7}
    \cmidrule(lr){8-10}
    \cmidrule(lr){11-13}
    \textbf{Condition}
        & Corr.
        & P. Corr.
        & Wrong
        & Corr.
        & P. Corr.
        & Wrong
        & Corr.
        & P. Corr.
        & Wrong
        & Corr.
        & P. Corr.
        & Wrong
    \\
    \cmidrule(lr){1-1}
    \cmidrule(lr){2-7}
    \cmidrule(lr){8-13}
    Increasing
        & 8 & 9 & 3 & 8 & 9 & 3
    \\
    Constant
        & 12 & 6 &  2 & 12 & 6 & 2 & 12 & 6 & 2 & 12 & 6 & 2
    \\
    Decreasing
        & 16 & 3 & 1 & 16 & 3 & 1
    \\
    \bottomrule
\end{tabular}
    }
\end{table*}


\paragraph{Procedure.} All participants provided informed consent and received identical instructions on how to perform the debugging task, without any mention of the different experimental conditions. To encourage attentiveness, we told participants that at the end of data collection, five debugging trials would be randomly
selected. Three participants, chosen at random from those who provided correct responses in all
five trials, would receive a £10 bonus. To ensure fairness, both random selections were carried
out, and the bonus was awarded accordingly. Participants  completed six warm-up trials to familiarize themselves with the task before starting the actual debugging session, which involved evaluating 40 images. All conditions presented a model with an overall accuracy equal to 60\%, but they differed in how the model's performance evolved throughout the debugging session. Between the first and the second half of the session, model's accuracy increased (from 40\% to 80\%) in the \textit{Inc} condition, remained constant (60\%) in the \textit{Const} condition, and decreased (from 80\% to 40\%) in the \textit{Dec} condition (see \cref{tab:exp1-conditions}). The order of presentation of the images was randomized once and kept fixed afterward for all participants within each condition. After the debugging task, participants filled out the questionnaire on perceived model’s accuracy and trustworthiness. Finally, they were asked to provide information about their experience with intelligent systems and debugging procedures.

\begin{figure}[!t]
    \centering
    \includegraphics[width=0.92\linewidth]{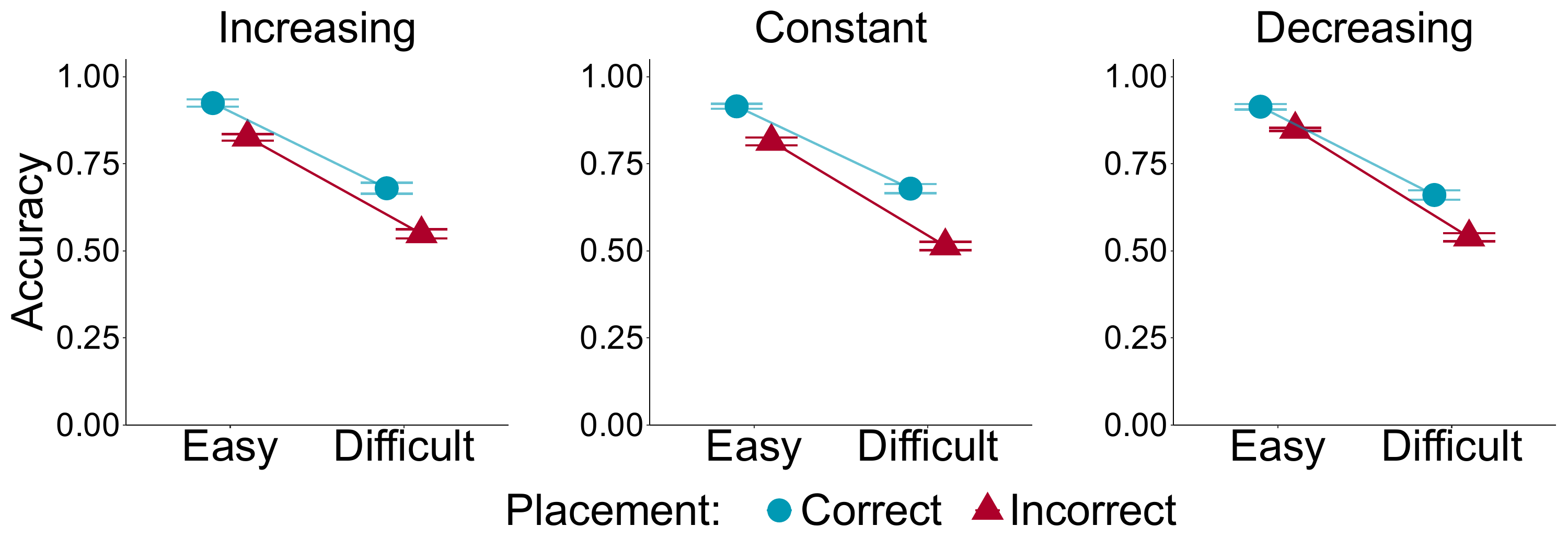}

    \vspace{0.5em}
    
    \includegraphics[width=0.92\linewidth]{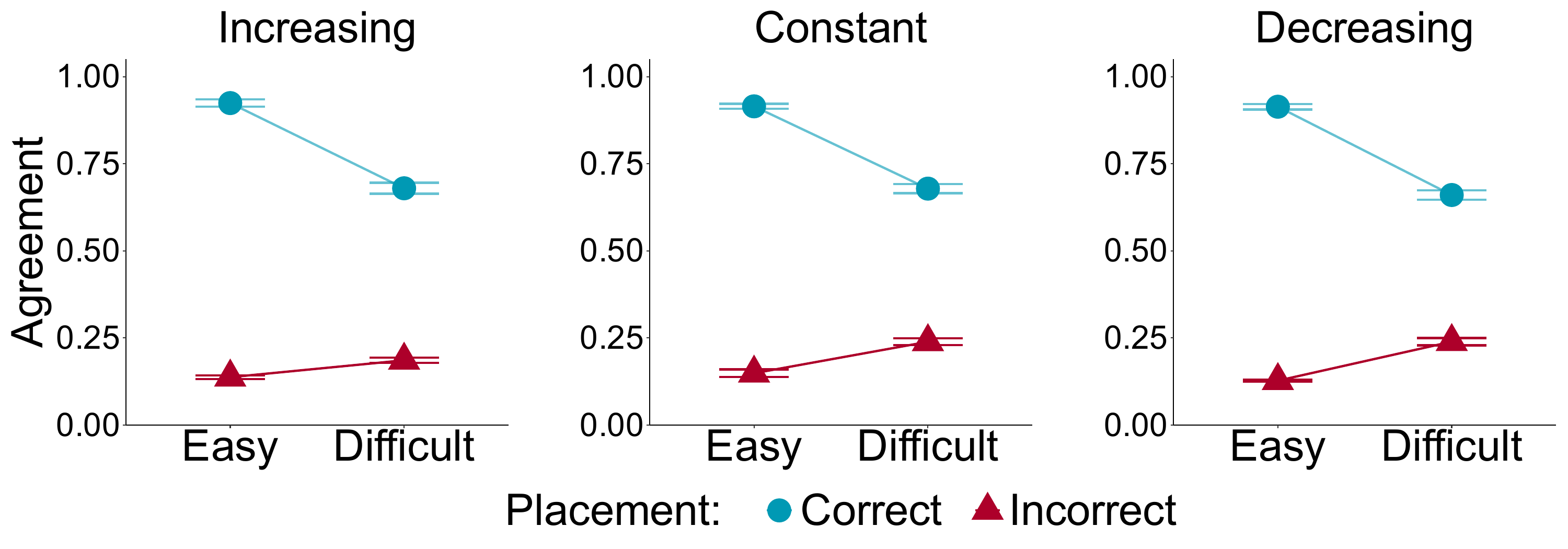}

    \caption{
        \textbf{Average accuracy} (top) and \textbf{agreement} (bottom) in Experiment 1 divided by order condition, correctness of model’s placement (correct and incorrect), and image difficulty. Error bars represent standard errors. 
    }
    \label{fig:exp1-results1}
\end{figure}


\paragraph{Sample composition and data quality.}
A total of 359 participants ($M_{Age} = $ $34.58 \pm 9.72$; $51\%$ female) were recruited and  evenly distributed across the three order conditions ($N_{\textit{Inc}} = 119$; $N_{\textit{Const}} = 121$; $N_{\textit{Dec}} = 119$).  A chi-square test of independence indicated that participants’ experience with programming and debugging machine learning algorithms did not differ significantly between the three conditions ($p = .188$), ensuring balanced samples in this regard. Also, the proportion of excluded observations was limited and roughly equal in the three order conditions ($\textit{Inc} = 8\%$; $\textit{Const} = 7\%$; $\textit{Dec} = 8\%$), suggesting that participants actively engaged with the task during the majority of trials.

\subsubsection{Results}

\paragraph{Accuracy.}
Participants' overall accuracy in the debugging task  was fairly high in all three order conditions (\textit{Inc}: $0.76 \pm 0.10$; \textit{Const}: $0.75\pm 0.08$; \textit{Dec}: $0.76 \pm 0.08$), albeit it varied depending on stimuli features: participants were more accurate when evaluating easy ($0.88 \pm 0.08$) than difficult images ($0.62 \pm 0.12$; $F(1, 36) = 63.33$, $p < .001$) and when evaluating correct ($0.80 \pm 0.11$) than incorrect images ($0.69 \pm 0.10$; $F(1, 36) = 10.65$, $p = .002$, see \cref{fig:exp1-results1}).\footnote{For significant effects of the mixed models we report the value of the test statistic ($F$) and relative degrees of freedom (in parentheses).}As for the effect of order condition, only its two-way interaction with model’s placement was significant, $F(2, 12875) = 4.56$, $p = .011$. Post-hoc tests revealed that the interaction was driven by participants' accuracy being more similar between incorrect and correct images in the \textit{Dec} condition (Correct: $0.79 \pm 0.11$; Incorrect: $0.70 \pm 0.08$) compared to the \textit{Const} one (Correct: $0.80 \pm 0.10$; Incorrect: $0.67 \pm 0.11$), $p = .008$. A possible interpretation of this result is that participants in the \textit{Dec} condition encountered most of the incorrect images in the latter half of the experiment, by which point they may have gained more confidence with the task. However, given that the difference was small and limited to the contrast between \textit{Dec} and \textit{Const}, it may also reflect random noise.

\paragraph{Agreement.}

 As expected, there was a main effect of the model’s placement, with agreement being higher when the box was placed correctly ($0.80 \pm 0.11$) than when it was not ($0.18 \pm 0.08$), $F(1, 36) = 174.38, p < .001$. The interaction between model's placement and image difficulty was significant as well, $F(1, 36) = 12.80$, $p = .001$, with agreement rates for difficult images being significantly lower than that for easy ones when the box was correctly placed  (Easy: $0.92 \pm 0.10$; Difficult: $0.67 \pm 0.16$, $p < .001$) but not when it was placed incorrectly (Easy: $0.14 \pm 0.08$; Difficult:  $0.22 \pm 0.11$, $p = .458$). Crucially, the three-way interaction was significant, $F(2, 12862) = 7.99$, $p < .001$:  when the box placement was incorrect, in all conditions, the agreement between participants and the model tended to be higher for difficult compared to easy images; however this tendency was reduced in the \textit{Inc} condition (Easy: $0.14 \pm 0.06$; Difficult: $0.19 \pm 0.08$) compared to the \textit{Const} (Easy: $0.15 \pm 0.12$; Difficult: $0.24 \pm 0.11$; $p = .027$) and the \textit{Dec} ones (Easy: $0.13 \pm 0.03$; Difficult: $0.24 \pm 0.12$; $p < .001$). This finding may be interpreted as evidence of a small primacy effect: \textit{early exposure to the model’s inaccuracy, as in the \textit{Inc} condition, led participants to rely less on the model, especially when the images were ambiguous}. In line with this interpretation, the effect described above was primarily driven (see \cref{fig:exp1-results4}) by difficult, partially wrong images, for which participants were likely most uncertain about the correctness of the model’s placement.

\paragraph{Questionnaire.}
The Kruskal-Wallis test on the perceived accuracy and trust index indicated that the three order conditions did not differ in how they evaluated the model after the debugging session ($p = .909$). Indeed, the average values of the index were almost identical for participants in the \textit{Inc} ($3.06 \pm 1.03$), \textit{Const} ($3.11 \pm 1.11$), and \textit{Dec} ($3.13 \pm 1.00$) conditions.


\paragraph{Discussion.}
The results of Experiment 1 suggest that the order in which participants are exposed to correct and incorrect explanations from a model within a single debugging session have, at best, a limited effect on the quality of their feedback and on their tendency to agree with the model. The only finding that is clearly interpretable as an order effect is that experiencing a consistent number of model failures early in a debugging session may reduce participants’ reliance on the model’s outputs, particularly when dealing with difficult stimuli. Interestingly, participants appeared to be unaware of this effect, as their responses to the questionnaire on perceived accuracy and trust showed no differences across order conditions.

\subsection{Experiment 2: Between-session Order Effect}
\label{sec:exp2}

\paragraph{Aim.}
In Experiment 2, we investigated possible order effects \textit{between two distinct sessions of a debugging task}. We presented participants with a simulated model whose accuracy either increased, remained constant, or decreased after a fictitious update between the two sessions, with the second session being identical for all participants.

\paragraph{Procedure.} 
The procedure was similar to that of Experiment 1. However, in this case, after the six warm-up trials, participants completed two debugging sessions, each consisting in the evaluation of 40 images. In the first session, the model’s accuracy varied across the three order conditions (\textit{Inc}: 40\%; \textit{Const}: 60\%; \textit{Dec}: 80\%).
The images were selected, based on the number required by each condition, from a larger set consisting of 32 correct images (16 easy and 16 difficult), 18 partially wrong images (9 easy and 9 difficult), and 6 wrong images (3 easy and 3 difficult). (See \cref{tab:exp1-conditions}.) This selection was carried out through 6 random draws, one for each image type. For example, the 8 correct easy images of the Inc condition were a subset of the 12 correct easy images of the Const condition, which, in turn, were a subset of the 16 correct easy images of the Dec condition.
These subsets were randomly drawn once, and the order of images was shuffled and kept constant across participants within each condition.

Following the first session, participants completed the same questionnaire used in Experiment 1. They then waited for five seconds, during which they were told the model was being updated based on feedback from all users. Participants then proceeded to the second session, which was identical across all order conditions in terms of both the images shown (model's accuracy was 60\%) and presentation order. Importantly, within both sessions, image presentation was controlled to distribute errors evenly in order to minimize potential within-session order effects. After completing the second session, participants, once again, filled in the same questionnaire and reported on their prior experiences with intelligent systems and debugging procedures.


\begin{figure}[!t]
    \centering
    \includegraphics[width=0.92\linewidth]{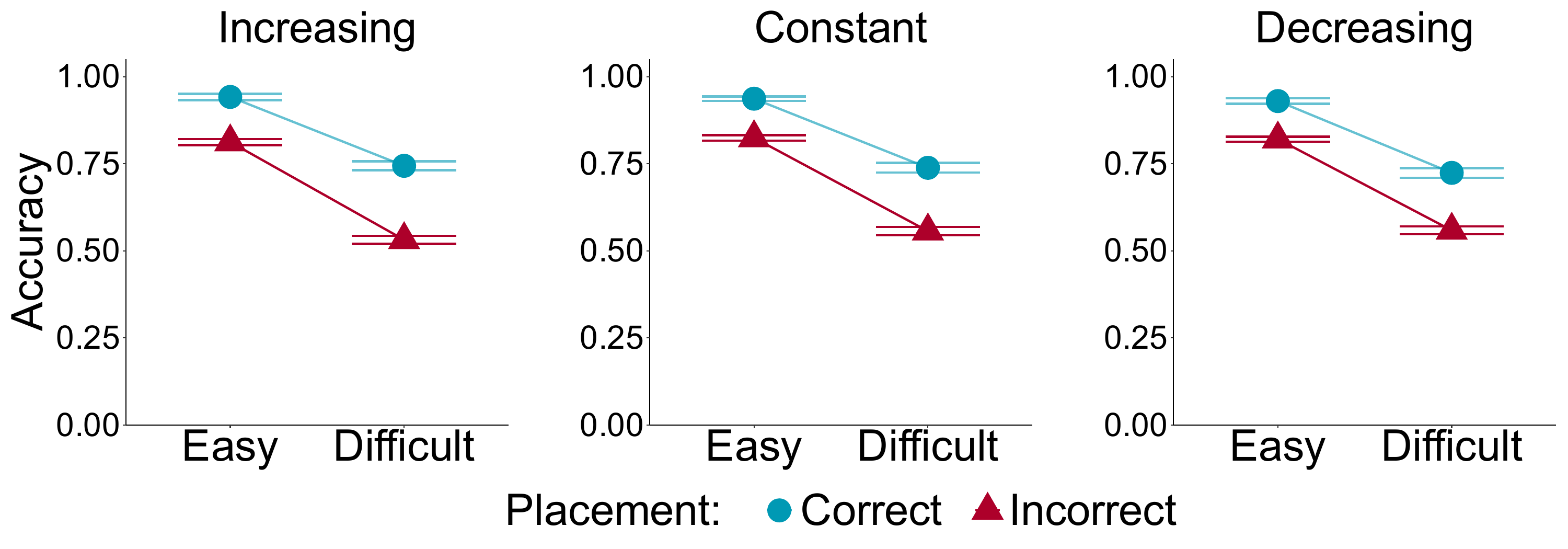}

    \vspace{0.5em}

    \includegraphics[width=0.92\linewidth]{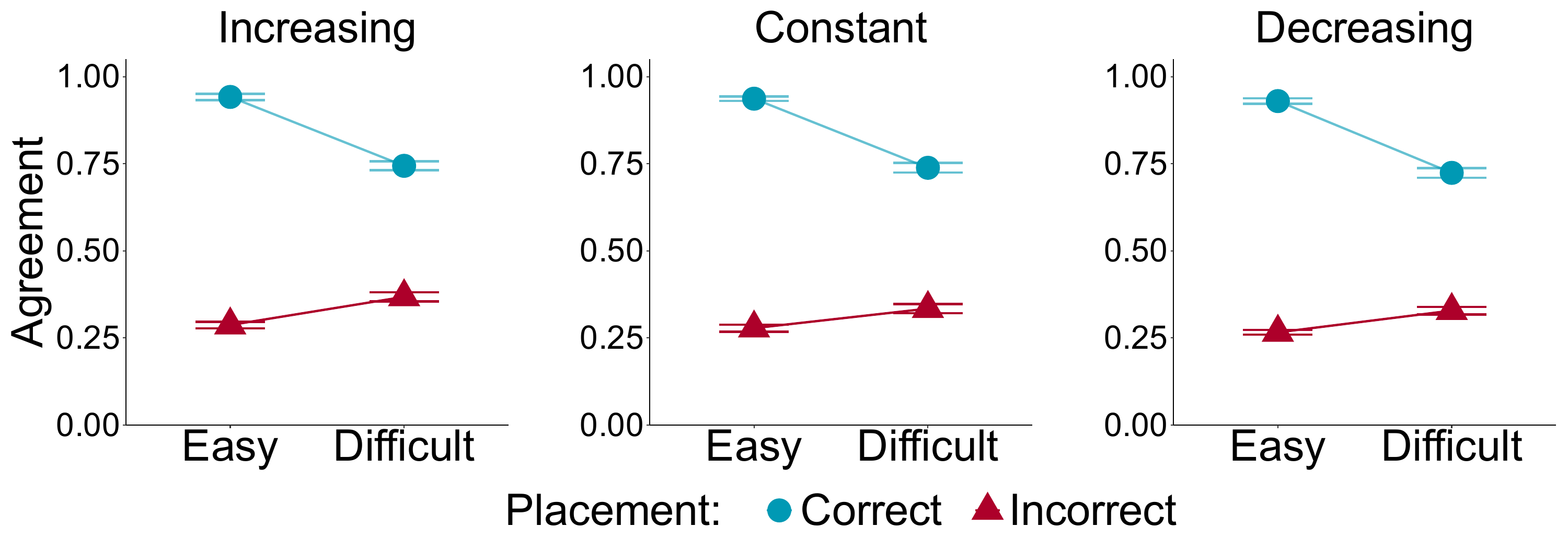}

    \caption{
        \textbf{Average accuracy} (top) and \textbf{agreement} (bottom) in Experiment 2 divided by order condition, correctness of model’s placement (correct and incorrect), and image difficulty. Error bars represent standard errors. 
    }
    \label{fig:exp2-results}
\end{figure}


\paragraph{Sample composition and data quality.}
A total of 354 participants ($M_{Age} = $ $36.33 \pm 10.64$, 47.5\% female) were recruited and randomly distributed across the three order conditions ($N_{\textit{Inc}} = 121$; $N_{\textit{Const}} = 117$; $N_{\textit{Dec}} = 116$). Yet again, experience with programming and debugging machine learning algorithms did not differ significantly between the three order conditions, $p = .278$, and engagement with the task was satisfactory, as indicated by the limited proportion of excluded observations $\textit{Inc} = 6\%$; $\textit{Const} = 6\%$; $\textit{Dec} = 7\%$).

\subsubsection{Results}

\paragraph{Manipulation check.}
To ensure that the manipulation of the model performance in the first session was effective, we assessed the agreement with the model through a mixed model including  order condition as a fixed effect and random intercepts for participants. Order condition predicted agreement, $F(2, 305.92) = 483.39$, $p < .001$, with post-hoc comparisons indicating that agreement was significantly lower in the \textit{Inc} ($0.42 \pm 0.36$) than in the \textit{Const} condition ($0.57 \pm 0.38$, $p < .001$) and in the \textit{Const} than in the \textit{Dec} condition ($0.70 \pm 0.35$, $p < .001$). 


\paragraph{Accuracy.}
In the second session, participants in all three order conditions showed a good average level of accuracy (\textit{Inc}: $0.78 \pm 0.08$; \textit{Const}: $0.78\pm 0.07$; \textit{Dec}: $0.78 \pm 0.08$). As expected, this was significantly higher for easy ($0.89 \pm 0.06$) than for difficult images ($0.66 \pm 0.11$; $F(1, 36) = 47.81$, $p < .001$) and for correct ($0.84 \pm 0.11$) than for incorrect images ($0.69 \pm 0.09$; $F(1, 36) = 19.55$, $p < .001$, see \cref{fig:exp2-results}, top). As in Experiment 1, the effect of order condition was significant only in interaction with model's placement, $F(2, 12915.6) = 8.61$, $p < .001$. Post-hoc contrasts indicated that, in this case, the effect was driven by a slightly more pronounced reduction in the accuracy for incorrect than correct images in the \textit{Inc} condition (Correct: $0.84 \pm 0.11$; Incorrect: $0.68 \pm 0.09$) compared to both the \textit{Const} (Correct: $0.84 \pm 0.10$; Incorrect: $0.69 \pm 0.09$; $p = .002$) and \textit{Dec} conditions (Correct: $0.83 \pm 0.11$; Incorrect: $0.70 \pm 0.09$; $p = .001$). One possible interpretation of this result is that participants in the Inc condition became more reliant on the model during the second session, following a perceived improvement in its accuracy. However, the small magnitude of the effect warrants caution in interpreting this finding, as it may simply reflect noise.

\paragraph{Agreement.}
In the second session, as in Experiment 1, we observed a significant main effect of model's placement on participants' agreement with the model, which was higher when the box was placed correctly ($0.84 \pm 0.11$) than when it was not ($0.31 \pm 0.10$; $F(1, 36) = 112.04$, $p < .001$). The effect was further qualified by a significant interaction with image difficulty ($F(1, 36) = 7.06$, $p = .012$): difficult images resulted in significantly lower agreement than easy ones when the box was correct (Easy: $0.94 \pm 0.08$; Difficult: $0.74 \pm 0.15$; $p = .006$), but not when it was incorrect (Easy: $0.28 \pm 0.10$; Difficult: $0.34 \pm 0.14$; $p = .816$). Crucially, no significant effect of order condition was observed, and indeed participants' overall agreement in the \textit{Inc} ($0.63 \pm 0.10$), \textit{Const} ($0.62 \pm 0.10$), and \textit{Dec} conditions ($0.61 \pm 0.09$) were similar and all converging toward intermediate values (see \cref{fig:exp2-results}, bottom).

\paragraph{Questionnaire.} As in Experiment 1, the Kruskal-Wallis test on the perceived accuracy and trust index suggested that participants in the \textit{Inc} ($3.38 \pm 1.21$), \textit{Const} ($3.43 \pm 1.17$), and \textit{Dec} ($3.34 \pm 1.23$) conditions did not differ in their perceptions of the model ($p = .821$).

\paragraph{Discussion.}
The results of Experiment 2 indicate that between-session order effects have a limited impact on users during XIL debugging sessions. Specifically, while there may be a minimal influence of order on users’ accuracy, the order of debugging sessions clearly did not affect participants’ agreement with the model or their perception of it.


%

\label{sec:related-work}

\section{Conclusion}
\label{sec:conclusion}

XIL can substantially help to enhance and steer the behavior of AI models. 
A number of XIL approaches have been proposed, see \citep{lertvittayakumjorn2021explanation, teso2023leveraging} for an overview.
While some methods are model-agnostic \citep{plumb2021finding, slany2022caipi, michiels2023increasing}, others are tailored for specific architectures, including neural networks \citep{teso2019toward, mitsuhara2019embedding, schramowski2020making, shao2021right} and concept-based models \citep{lertvittayakumjorn2020find, stammer2021right}.
However, \textit{most works focus on algorithmic, rather than human, factors}, such as how to best integrate feedback into the model \citep{schramowski2020making, shao2021right, michiels2023increasing}, and how to leverage alternatives to saliency maps, like examples \citep{teso2021interactive, zylberajch2021hildif} and high-level concepts \citep{stammer2021right}. In XIL, items are choosen by the machine, usually according to their predictive uncertainty \citep{settles2011closing}, and presented sequentially, potentially triggering order effects.
\citet{popordanoska2020machine} argue that this setup prioritizes easy-to-learn items, and as such it may fool users into overtrusting the model, and propose human-initiated interaction as a possible solution \citep{attenberg2010label}.  Their observation that the order in which items are presented -- which depends entirely on the machine's query selection strategy -- can induce overreliance partially motivates our work.

\paragraph{Comparison with Related Work.}
Previous results suggested that letting users interact with a model during debugging can unduly decrease their trust in it \citep{honeycutt2020soliciting} and expose them to the influence of biases, worsening the reliability of their feedback (\citet{nourani2021anchoring}).
We examined these potential issues by engaging users in a realistic XIL debugging task and assessing whether participants' performance and perceptions were influenced by within- and between-session order effects, which may be both naturally triggered by the XIL loop.
Our results showed that participants provided high-quality feedback and appropriately adjusted their agreement with the model based on its performance, and that these variables were only minimally affected by order effects. These results were confirmed by participants' self-reported perceptions, which did not show any difference between conditions In particular, we observed that participants' agreement with the model was influenced by presentation order only within (but not between) sessions, and only with certain kind of stimuli (i.e. difficult images), suggesting that participants may reset their expectations when the model is updated. This possibility is reassuring, since it indicates that participants can adapt to model changes without being biased by prior exposure, making  XIL algorithms overall robust to the influence of order effects, provided users are informed about model updates. This finding warrants further investigation to determine the optimal procedure for implementing debugging sessions.

\paragraph{Limitations and future directions.}
Our work focuses exclusively on order effects arising from differences in the distribution of model errors, which are naturally triggered by the XIL loop. Future work should consider other types of prediction tasks, beyond image classification, and other widely used families of explanations, such as concept-level explanations \citep{stammer2021right} and concrete examples \citep{zylberajch2021hildif}. It could also be worth exploring different types of order effects, \textit{e.g.} those stemming from the sequence in which more or less challenging items or explanations are presented, as these may influence participants’ accuracy by modulating their confidence in their ability to perform the task.
More broadly, future work could examine whether the implementation of XIL algorithms is effective across different types of tasks, and whether other forms of bias might more substantially hinder their adoption.

\newpage

\section*{Ethical Statement}
Our study has received approval from the Ethics board of our university, document identifier code 2025-001ESA.

\section*{Acknowledgments}
Funded by the European Union. Views and opinions expressed are however those of the author(s) only and do not necessarily reflect those of the European Union or the European Health and Digital Executive Agency (HaDEA). Neither the European Union nor the granting authority can be held responsible for them. Grant Agreement no. 101120763 - TANGO.

\bibliographystyle{plainnat}
\bibliography{paper,explanatory-supervision}

@inproceedings{ross2017right,
  title={Right for the right reasons: training differentiable models by constraining their explanations},
  author={Ross, Andrew Slavin and Hughes, Michael C and Doshi-Velez, Finale},
  booktitle={Proceedings of the 26th International Joint Conference on Artificial Intelligence},
  pages={2662--2670},
  year={2017}
}

@article{lertvittayakumjorn2021explanation,
  title={Explanation-Based Human Debugging of NLP Models: A Survey},
  author={Lertvittayakumjorn, Piyawat and Toni, Francesca},
  journal={arXiv preprint arXiv:2104.15135},
  year={2021}
}

@inproceedings{kulesza2015principles,
  title={Principles of explanatory debugging to personalize interactive machine learning},
  author={Kulesza, Todd and Burnett, Margaret and Wong, Weng-Keen and Stumpf, Simone},
  booktitle={Proceedings of the 20th international conference on intelligent user interfaces},
  pages={126--137},
  year={2015}
}

@inproceedings{teso2019explanatory,
  title={Explanatory interactive machine learning},
  author={Teso, Stefano and Kersting, Kristian},
  booktitle={Proceedings of the 2019 AAAI/ACM Conference on AI, Ethics, and Society},
  pages={239--245},
  year={2019}
}

@inproceedings{teso2019toward,
  title={Toward Faithful Explanatory Active Learning with Self-explainable Neural Nets},
  author={Teso, Stefano},
  booktitle={Proceedings of the Workshop on Interactive Adaptive Learning (IAL 2019)},
  pages={4--16},
  year={2019}
}

@article{schramowski2020making,
  title={Making deep neural networks right for the right scientific reasons by interacting with their explanations},
  author={Schramowski, Patrick and Stammer, Wolfgang and Teso, Stefano and Brugger, Anna and Herbert, Franziska and Shao, Xiaoting and Luigs, Hans-Georg and Mahlein, Anne-Katrin and Kersting, Kristian},
  journal={Nature Machine Intelligence},
  volume={2},
  number={8},
  pages={476--486},
  year={2020},
  publisher={Nature Publishing Group}
}

@inproceedings{honeycutt2020soliciting,
  title={Soliciting human-in-the-loop user feedback for interactive machine learning reduces user trust and impressions of model accuracy},
  author={Honeycutt, Donald and Nourani, Mahsan and Ragan, Eric},
  booktitle={Proceedings of the AAAI Conference on Human Computation and Crowdsourcing},
  volume={8},
  number={1},
  pages={63--72},
  year={2020}
}

@article{mitsuhara2019embedding,
  title={Embedding Human Knowledge into Deep Neural Network via Attention Map},
  author={Mitsuhara, Masahiro and Fukui, Hiroshi and Sakashita, Yusuke and Ogata, Takanori and Hirakawa, Tsubasa and Yamashita, Takayoshi and Fujiyoshi, Hironobu},
  journal={arXiv preprint arXiv:1905.03540},
  year={2019}
}

@inproceedings{lertvittayakumjorn2020find,
  title={FIND: human-in-the-loop debugging deep text classifiers},
  author={Lertvittayakumjorn, Piyawat and Specia, Lucia and Toni, Francesca},
  booktitle={Conference on Empirical Methods in Natural Language Processing},
  pages={332--348},
  year={2020}
}

@article{popordanoska2020machine,
  title={{Machine Guides, Human Supervises: Interactive Learning with Global Explanations}},
  author={Popordanoska, Teodora and Kumar, Mohit and Teso, Stefano},
  journal={arXiv preprint arXiv:2009.09723},
  year={2020}
}

@inproceedings{stammer2021right,
  title={{Right for the Right Concept: Revising Neuro-Symbolic Concepts by Interacting with their Explanations}},
  author={Stammer, Wolfgang and Schramowski, Patrick and Kersting, Kristian},
  booktitle={Proceedings of the IEEE/CVF Conference on Computer Vision and Pattern Recognition},
  pages={3619--3629},
  year={2021}
}

@inproceedings{shao2021right,
  title={{Right for Better Reasons: Training Differentiable Models by Constraining their Influence Function}},
  author={Shao, Xiaoting and Skryagin, Arseny and Schramowski, P and Stammer, W and Kersting, Kristian},
  booktitle={Proceedings of Thirty-Fifth AAAI Conference on Artificial Intelligence (AAAI)},
  year={2021}
}

@article{zylberajch2021hildif,
  title={{HILDIF: Interactive Debugging of NLI Models Using Influence Functions}},
  author={Zylberajch, Hugo and Lertvittayakumjorn, Piyawat and Toni, Francesca},
  journal={Workshop on Interactive Learning for Natural Language Processing},
  pages={1},
  year={2021}
}

@inproceedings{teso2021interactive,
  title={{Interactive Label Cleaning with Example-based Explanations}},
  author={Teso, Stefano and Bontempelli, Andrea and Giunchiglia, Fausto and Passerini, Andrea},
  booktitle={Proceedings of the 35th International Conference on Neural Information Processing Systems},
  year={2021}
}

@article{plumb2021finding,
  title={{Finding and Fixing Spurious Patterns with Explanations}},
  author={Plumb, Gregory and Ribeiro, Marco Tulio and Talwalkar, Ameet},
  journal={arXiv preprint arXiv:2106.02112},
  year={2021}
}

@inproceedings{slany2022caipi,
  title={CAIPI in Practice: Towards Explainable Interactive Medical Image Classification},
  author={Slany, Emanuel and Ott, Yannik and Scheele, Stephan and Paulus, Jan and Schmid, Ute},
  booktitle={IFIP International Conference on Artificial Intelligence Applications and Innovations},
  pages={389--400},
  year={2022},
  organization={Springer}
}

@article{teso2023leveraging,
  title={Leveraging Explanations in Interactive Machine Learning: An Overview},
  author={Teso, Stefano and Alkan, {\"O}znur and Stammer, Wolfang and Daly, Elizabeth},
  journal={Frontiers in Artificial Intelligence},
  year={2023}
}

@inproceedings{settles2011closing,
  title={Closing the loop: Fast, interactive semi-supervised annotation with queries on features and instances},
  author={Settles, Burr},
  booktitle={Proceedings of the 2011 Conference on Empirical Methods in Natural Language Processing},
  pages={1467--1478},
  year={2011}
}

@article{miller2019explanation,
  title={Explanation in artificial intelligence: Insights from the social sciences},
  author={Miller, Tim},
  journal={Artificial intelligence},
  volume={267},
  pages={1--38},
  year={2019},
  publisher={Elsevier}
}

@article{lapuschkin2019unmasking,
  title={Unmasking Clever Hans predictors and assessing what machines really learn},
  author={Lapuschkin, Sebastian and W{\"a}ldchen, Stephan and Binder, Alexander and Montavon, Gr{\'e}goire and Samek, Wojciech and M{\"u}ller, Klaus-Robert},
  journal={Nature communications},
  volume={10},
  number={1},
  pages={1--8},
  year={2019},
  publisher={Nature Publishing Group}
}

@article{geirhos2020shortcut,
  title={Shortcut learning in deep neural networks},
  author={Geirhos, Robert and Jacobsen, J{\"o}rn-Henrik and Michaelis, Claudio and Zemel, Richard and Brendel, Wieland and Bethge, Matthias and Wichmann, Felix A},
  journal={Nature Machine Intelligence},
  volume={2},
  number={11},
  pages={665--673},
  year={2020},
  publisher={Nature Publishing Group}
}

@inproceedings{attenberg2010label,
  title={Why label when you can search? Alternatives to active learning for applying human resources to build classification models under extreme class imbalance},
  author={Attenberg, Josh and Provost, Foster},
  booktitle={Proceedings of the 16th ACM SIGKDD international conference on Knowledge discovery and data mining},
  pages={423--432},
  year={2010}
}

@book{molnar2020interpretable,
  title={Interpretable machine learning},
  author={Molnar, Christoph},
  year={2020},
  publisher={Lulu. com}
}

@article{schwalbe2024comprehensive,
  title={A comprehensive taxonomy for explainable artificial intelligence: a systematic survey of surveys on methods and concepts},
  author={Schwalbe, Gesina and Finzel, Bettina},
  journal={Data Mining and Knowledge Discovery},
  volume={38},
  number={5},
  pages={3043--3101},
  year={2024},
  publisher={Springer}
}

@article{eisenberg1988order,
  title={Order effects: A study of the possible influence of presentation order on user judgments of document relevance},
  author={Eisenberg, Michael and Barry, Carol},
  journal={Journal of the American Society for Information Science},
  volume={39},
  number={5},
  pages={293--300},
  year={1988},
  publisher={Wiley Online Library}
}

@article{matsusaka2016ballot,
  title={Ballot order effects in direct democracy elections},
  author={Matsusaka, John G},
  journal={Public choice},
  volume={167},
  pages={257--276},
  year={2016},
  publisher={Springer}
}

@article{rastogi2022deciding,
  author = {Rastogi, Charvi and others},
  title = {Deciding Fast and Slow: The Role of Cognitive Biases in AI-Assisted Decision-Making},
  year = {2022},
  journal = {Proc. ACM Hum.-Comput. Interact.},
}

@inproceedings{cummings2012,
author = {Mary Cummings},
title = {Automation Bias in Intelligent Time Critical Decision Support Systems},
booktitle = {Collection of Technical Papers - AIAA 1st Intelligent Systems Technical Conference},
year = 2012
}

@misc{hoffman2019metrics,
      title={Metrics for Explainable AI: Challenges and Prospects}, 
      author={Robert R. Hoffman and Shane T. Mueller and Gary Klein and Jordan Litman},
      year={2019},
      eprint={1812.04608},
      archivePrefix={arXiv},
      primaryClass={cs.AI}
}

@inproceedings{nourani2020role,
  title={The role of domain expertise in user trust and the impact of first impressions with intelligent systems},
  author={Nourani, Mahsan and King, Joanie and Ragan, Eric},
  booktitle={Proceedings of the AAAI Conference on Human Computation and Crowdsourcing},
  volume={8},
  pages={112--121},
  year={2020}
}

@inproceedings{nourani2021anchoring,
  title={Anchoring bias affects mental model formation and user reliance in explainable ai systems},
  author={Nourani, Mahsan and Roy, Chiradeep and Block, Jeremy E and Honeycutt, Donald R and Rahman, Tahrima and Ragan, Eric and Gogate, Vibhav},
  booktitle={26th International Conference on Intelligent User Interfaces},
  pages={340--350},
  year={2021}
}

@article{FURNHAM201135,
title = {A literature review of the anchoring effect},
journal = {The Journal of Socio-Economics},
volume = {40},
number = {1},
pages = {35-42},
year = {2011},
issn = {1053-5357},
doi = {https://doi.org/10.1016/j.socec.2010.10.008},
url = {https://www.sciencedirect.com/science/article/pii/S1053535710001411},
author = {Adrian Furnham and Hua Chu Boo}
}

@article{baddeley1993recency,
  title={The recency effect: Implicit learning with explicit retrieval?},
  author={Baddeley, Alan D and Hitch, Graham},
  journal={Memory \& Cognition},
  volume={21},
  pages={146--155},
  year={1993},
  publisher={Springer}
}

@article{HOGARTH19921,
title = {Order effects in belief updating: The belief-adjustment model},
journal = {Cognitive Psychology},
volume = {24},
number = {1},
pages = {1-55},
year = {1992},
issn = {0010-0285},
doi = {https://doi.org/10.1016/0010-0285(92)90002-J},
url = {https://www.sciencedirect.com/science/article/pii/001002859290002J},
author = {Robin M Hogarth and Hillel J Einhorn}}

@article{nourani2022importance,
  title={On the importance of user backgrounds and impressions: Lessons learned from interactive AI applications},
  author={Nourani, Mahsan and Roy, Chiradeep and Block, Jeremy E and Honeycutt, Donald R and Rahman, Tahrima and Ragan, Eric D and Gogate, Vibhav},
  journal={ACM Transactions on Interactive Intelligent Systems},
  volume={12},
  number={4},
  pages={1--29},
  year={2022},
  publisher={ACM New York, NY}
}

@article{green2016simr,
  title={SIMR: An R package for power analysis of generalized linear mixed models by simulation},
  author={Green, Peter and MacLeod, Catriona J},
  journal={Methods in Ecology and Evolution},
  volume={7},
  number={4},
  pages={493--498},
  year={2016}
}

@article{kumle2021estimating,
  title={Estimating power in (generalized) linear mixed models: An open introduction and tutorial in R},
  author={Kumle, Levi and V{\~o}, Melissa L-H and Draschkow, Dejan},
  journal={Behavior research methods},
  volume={53},
  number={6},
  pages={2528--2543},
  year={2021},
  publisher={Springer}
}

@article{michiels2023increasing,
  title={Increasing Performance And Sample Efficiency With Model-agnostic Interactive Feature Attributions},
  author={Michiels, Joran and De Vos, Maarten and Suykens, Johan},
  journal={arXiv preprint arXiv:2306.16431},
  year={2023}
}

@article{OpenImages,
  author = {Alina Kuznetsova and Hassan Rom and Neil Alldrin and Jasper Uijlings and Ivan Krasin and Jordi Pont-Tuset and Shahab Kamali and Stefan Popov and Matteo Malloci and Alexander Kolesnikov and Tom Duerig and Vittorio Ferrari},
  title = {The Open Images Dataset V4: Unified image classification, object detection, and visual relationship detection at scale},
  year = {2020},
  journal = {IJCV}
}

@article{ye2024spurious,
  title={Spurious correlations in machine learning: A survey},
  author={Ye, Wenqian and Zheng, Guangtao and Cao, Xu and Ma, Yunsheng and Zhang, Aidong},
  journal={arXiv preprint arXiv:2402.12715},
  year={2024}
}

@article{sheeran2016intention,
  title={The intention--behavior gap},
  author={Sheeran, Paschal and Webb, Thomas L},
  journal={Social and personality psychology compass},
  volume={10},
  number={9},
  pages={503--518},
  year={2016},
  publisher={Wiley Online Library}
}

@article{warren2024categorical,
  title={Categorical and continuous features in counterfactual explanations of AI systems},
  author={Warren, Greta and Byrne, Ruth MJ and Keane, Mark T},
  journal={ACM Transactions on Interactive Intelligent Systems},
  volume={14},
  number={4},
  pages={1--37},
  year={2024},
  publisher={ACM New York, NY}
}

\newpage
\appendix

\section{Experimental Details}

\subsection{Overlap formulas}
Below are reported the formulas used to compute the two behavioral dependent variables considered in Experiments 1 and 2 (participants' feedback accuracy  their agreement with the model) and a third metric, representing the accuracy of the model's explanation, based on which images were categorized as correct or incorrect. 

Please note that, within each image, the boxes enclosing the ground truth ($box_{GT}$), the model's explanation ($box_{Model}$), and the user final feedback ($box_{Part}$) were all equal in size. Thus, each metric returns a value between 0 and 1, corresponding to the proportion of overlap (\textit{i.e.} intersection) between two boxes of interest.

\begin{itemize}
    \item Accuracy of participant's feedback: $(box_{GT} \cap box_{Part}) / Area_{GT}$
    \item Participant's agreement with the model: $(box_{Model} \cap box_{User}) / Area_{Model}$
    \item Accuracy of model's explanation: $(box_{GT} \cap box_{Model}) / Area_{GT}$
\end{itemize}

\subsection{Generating images with wrong explanations}

The images presenting different levels of correctness of the model's box placement were generated as follows: correct images presented a model's accuracy of 1, partially wrong images an accuracy of 0.25, and wrong images an accuracy of 0. Box placement for the two types of incorrect images were performed manually to achieve the desired level of accuracy while maintaining the box within the image borders. 

\subsection{Pilot studies on image difficulty}

Images were categorised as easy or difficult based on the results of two pilot studies conducted prior to the main experiments. Participants in both pilots were recruited via Prolific using the same inclusion criteria as in the actual experiments. In each pilot, participants completed the same debugging task used in the main studies but evaluated a larger set of images than those ultimately included. From these larger sets, easy images were selected from those with the highest participant accuracy, while difficult images were drawn from those with the lowest accuracy.

The first pilot study (\textit{N} = 19) was used to select the 40 images employed in the single debugging session of Experiment 1. The second pilot study (\textit{N} = 20) was used to select the 56 images shown in the first session of Experiment 2. The second session of Experiment 2 employed the same 40 images from Experiment 1, with the exception of four wrong images (two easy and two difficult) in which the box placement was modified to become partially wrong (this decision was motivated by the fact that, in Experiment 1, order effects appeared most evident in trials involving partially wrong images). The difficulty of these four images was tested within the second pilot, which confirmed their original difficulty categorization.

Overall, the pilot studies enabled the selection of easy and difficult images that, within the same model placement category (\textit{i.e.} correct, partially wrong, or wrong), differed substantially in terms of participants’ average accuracy (see \ref{tab:within pilot} and \ref{tab:between pilot}).

\begin{table}[h!]
\centering
\begin{tabular}{ccc}
\toprule
\textbf{} & \textbf{Easy} & \textbf{Difficult} \\
\midrule
Correct & $0.98 \pm 0.02$ & $0.71 \pm 0.17$ \\
Partially wrong & $0.83 \pm 0.04$ & $0.41 \pm 0.05$ \\
Wrong & $0.85 \pm 0.03$ & $0.63 \pm 0.05$ \\
\bottomrule
\end{tabular}
\caption{Average accuracy per image category in the first pilot study.}
\label{tab:within pilot}
\end{table}

\begin{table}[h!]
\centering
\begin{tabular}{ccc}
\toprule
\textbf{} & \textbf{Easy} & \textbf{Difficult} \\
\midrule
Correct & $0.89 \pm 0.04$ & $0.75 \pm 0.08$ \\
Partially wrong & $0.82 \pm 0.04$ & $0.65 \pm 0.06$ \\
Wrong & $0.79 \pm 0.04$ & $0.64 \pm 0.06$ \\
\bottomrule
\end{tabular}
\caption{Average accuracy per image category in the second pilot study.}
\label{tab:between pilot}
\end{table}

\subsection{Questionnaire Items}
\label{sec:questionnaire-items}
\textbf{Trust and perception of accuracy}

The following questionnaire, adapted from \citet{hoffman2019metrics}, was administered following each debugging session. \\

\textbf{Answer each sentence with a number from 1 (completely disagree) to 7 (completely agree)}
\begin{itemize}
    \item I am confident in the model. I feel that it works well.
    \item The outputs of the model are very predictable.
    \item The model is very reliable. 
    \item The model can perform the task better than a novice human user.
\end{itemize}
\textbf{Programming Experience}

The following multiple-choice question was asked at the end of the experiment, \textit{i.e.}, after the sole debugging session in the within-session experiment and after the second session in the between-session experiment.\\

\textit{What is your experience with programming and machine learning algorithm debugging?}

\begin{itemize}
    \item I've never coded any kind of program.
    \item I've never coded any machine learning algorithm.
    \item I have limited experience (\textit{e.g.} university courses, online courses) with machine learning.
    \item I frequently use machine learning models but I have no experience in debugging them.
    \item I frequently use machine learning models and I have experience in debugging them.
    \item I frequently develop machine learning models and I have experience in debugging them.
\end{itemize}

\subsection{Instructions}

\subsubsection{Experiment 1 instructions} \hfill\\

Welcome to this study.

In what follows, you will see the outputs of an artificial intelligence model trained to
IDENTIFY THE LOCATION OF A HUMAN FACE in a blurred image,
BY PLACING A BOX AROUND IT.

The box should be placed only around the face,
excluding other parts of the body or clothing, regardless of their size.

Your task is to evaluate the model’s accuracy.
To this end, you will be shown 40 images.
Every image will contain a single human face and the box positioned by the model.

------------

Each image will be displayed for 6 seconds.

If you think the BOX could be BETTER POSITIONED (even slightly),
then CLICK WITH THE MOUSE ON THE POSITION YOU BELIEVE IS CORRECT.
Note: you can adjust the position as many times as you like within the 6-second time limit.

If you think the BOX HAS BEEN POSITIONED EXACTLY at the center of the face by the model,
then CLICK THE CONFIRMATION BUTTON WITH THE MOUSE.
Note: once you press this button, you will no longer be able to adjust the box’s position.

------------

We ask you to be as ACCURATE AS POSSIBLE.
We will also measure the time you take to respond,
but ALWAYS PRIORITIZE ACCURACY OVER SPEED.

Note: if you neither move the box nor confirm its position within the 6-second window,
your response for that image will be recorded as missing.

Before starting the actual task, you will be shown 6 practice images.
Just like in the main task, you can either click on the position you believe is correct,
or confirm the box placement made by the model.

------------

At the end of the session, you will be asked to complete a short questionnaire.

BONUS:
In addition to the payment for your participation, you have the chance to win an extra £10.
At the end of the experiment, five images will be randomly selected.
Among those who gave the correct answer for all five of these images,
three participants will be randomly chosen to receive a £10 bonus.

------------

You have the right to withdraw from the experiment at any time, without giving a reason for your withdrawal.
To do so, press the ‘ESC’ key. The experiment will be immediately terminated, and you will not receive any payment.

IMPORTANT:
Exit full-screen mode only if you intend to withdraw from the experiment.
If you exit full-screen mode before the experiment is completed, it will be interrupted and you will be considered as having withdrawn from the study.
As a result, you will not be paid (regardless of how much of the task you had completed).

By clicking this box,
I confirm that I have read and understood the information provided about the study,
and that I am participating voluntarily.

\hfill\\
\subsubsection{Experiment 2 instructions} \hfill\\
\label{between instr}

Welcome to this study.

In what follows, you will see the outputs of an artificial intelligence model trained to
IDENTIFY THE LOCATION OF A HUMAN FACE in a blurred image,
BY PLACING A BOX AROUND IT.

The box should be placed only around the face,
excluding other parts of the body or clothing, regardless of their size.

Your task is to evaluate the model’s accuracy.
To this end, you will be shown two sessions of 40 images each.
Every image will contain a single human face and the box positioned by the model.

There will be a short break between the two sessions,
to allow the model to be updated based on the feedback received from various users during the first session.

------------

Each image will be displayed for 6 seconds.

If you think the BOX could be BETTER POSITIONED (even slightly),
then CLICK WITH THE MOUSE ON THE POSITION YOU BELIEVE IS CORRECT.
Note: you can adjust the position as many times as you like within the 6-second time limit.

If you think the BOX HAS BEEN POSITIONED EXACTLY at the center of the face by the model,
then CLICK THE CONFIRMATION BUTTON WITH THE MOUSE.
Note: once you press this button, you will no longer be able to adjust the box’s position.

------------

We ask you to be as ACCURATE AS POSSIBLE.
We will also measure the time you take to respond,
but ALWAYS PRIORITIZE ACCURACY OVER SPEED.

Note: if you neither move the box nor confirm its position within the 6-second window,
your response for that image will be recorded as missing.

Before starting the actual task, you will be shown 6 practice images.
Just like in the main task, you can either click on the position you believe is correct,
or confirm the box placement made by the model.

------------

At the end of each session, you will be asked to complete a short questionnaire.

BONUS:
In addition to the payment for your participation, you have the chance to win an extra £10.
At the end of the experiment, five images will be randomly selected.
Among those who gave the correct answer for all five of these images,
three participants will be randomly chosen to receive a £10 bonus.

------------

You have the right to withdraw from the experiment at any time, without giving a reason for your withdrawal.
To do so, press the ‘ESC’ key. The experiment will be immediately terminated, and you will not receive any payment.

IMPORTANT:
Exit full-screen mode only if you intend to withdraw from the experiment.
If you exit full-screen mode before the experiment is completed, it will be interrupted and you will be considered as having withdrawn from the study.
As a result, you will not be paid (regardless of how much of the task you had completed).

By clicking this box,
I confirm that I have read and understood the information provided about the study,
and that I am participating voluntarily.



\newpage
\section{Supplementary images}
\label{sec:other-results}
In this section, we include additional plots to illustrate differences in participants' accuracy and their agreement with the model for images featuring a box whose placement was correct, partially wrong, and wrong.

\begin{figure*}[!h]
    \label{fig:exp1-results3}
    \centering
    \includegraphics[width=0.7\linewidth]{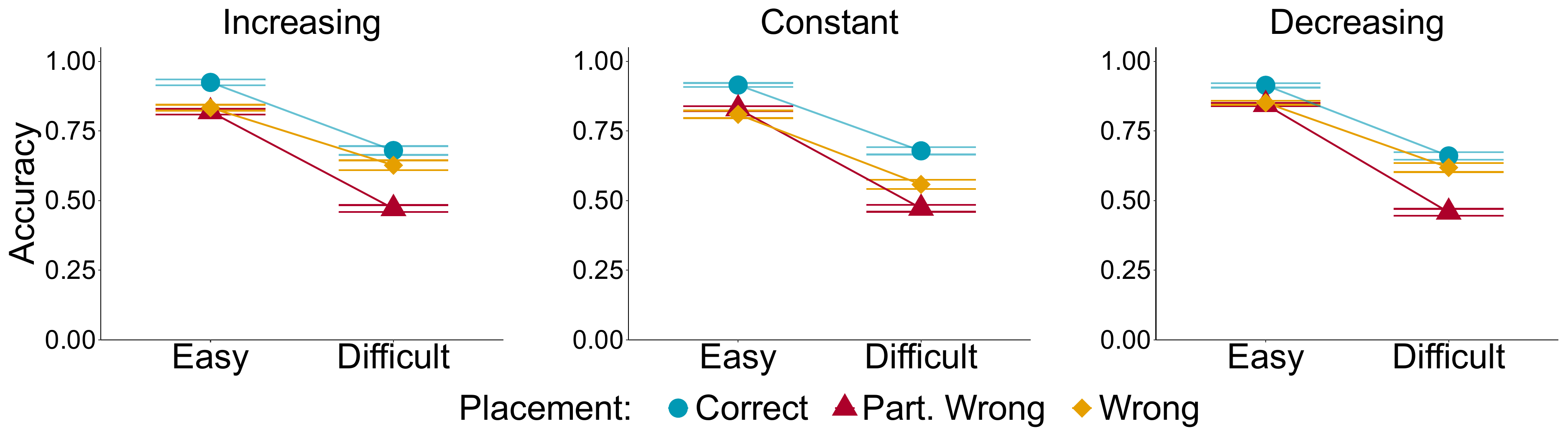}
        \caption{
        \textbf{Average accuracy} in Experiment 1 divided by order condition, correctness of model’s placement (correct, partially wrong, and wrong), and image difficulty. Error bars represent standard errors.
    }
\end{figure*}

\begin{figure*}[!h]
    \label{fig:exp1-results4}
    \centering
    \includegraphics[width=0.7\linewidth]{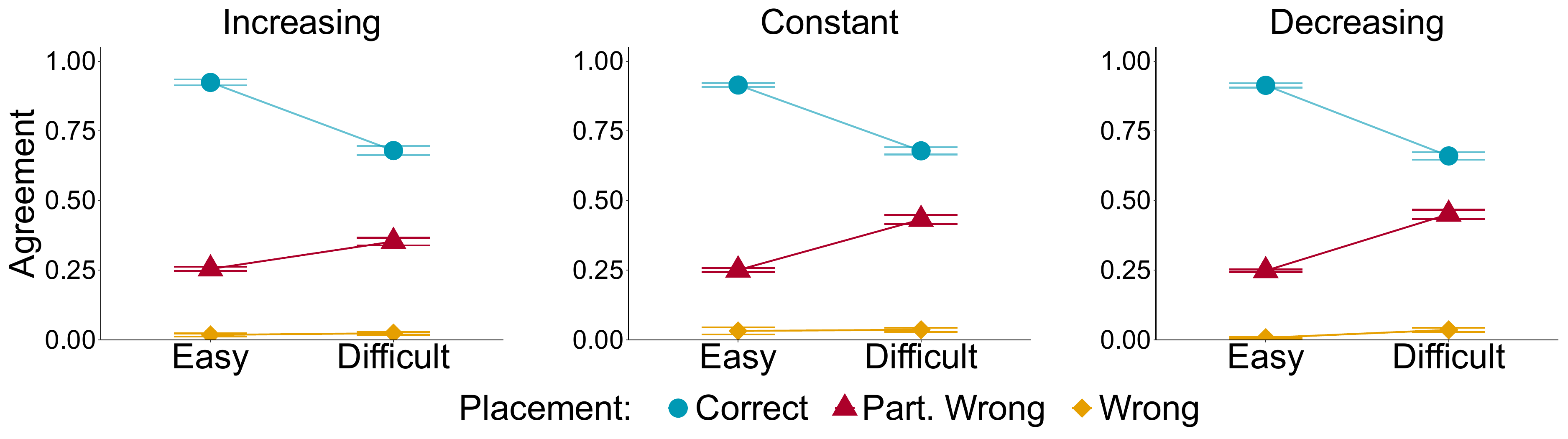}
        \caption{
        \textbf{Average agreement} in Experiment 1 divided by order condition, correctness of model’s placement (correct, partially wrong, and wrong), and image difficulty. Error bars represent standard errors.
    }
\end{figure*}

\begin{figure*}[!h]
    \label{fig:exp2-results3}
    \centering
    \includegraphics[width=0.7\linewidth]{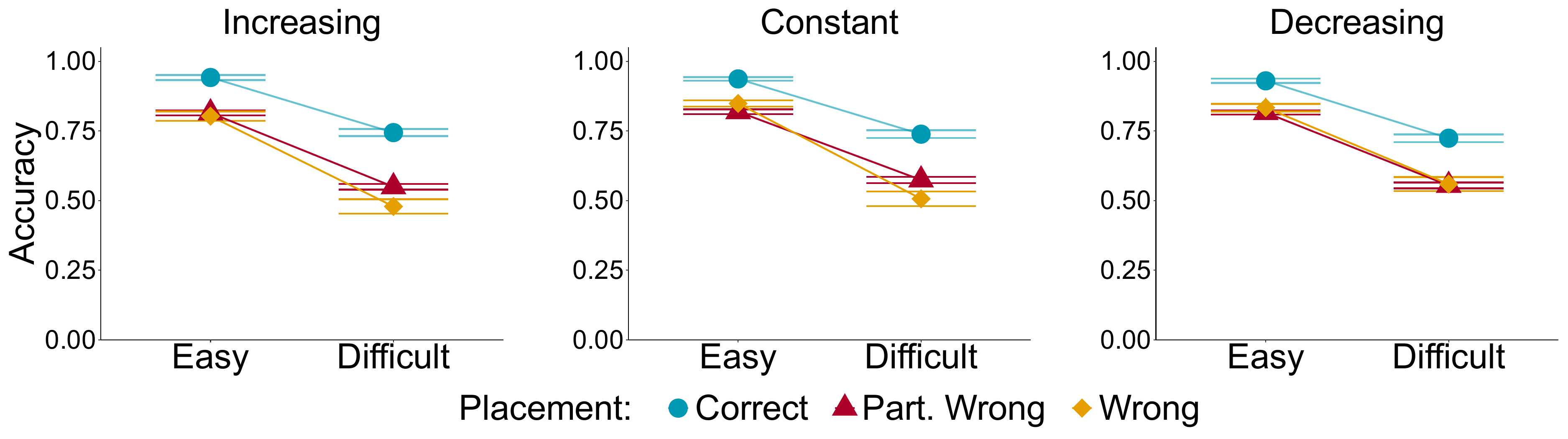}
        \caption{
        \textbf{Average accuracy} in Experiment 2 divided by order condition, correctness of model’s placement (correct, partially wrong, and wrong), and image difficulty. Error bars represent standard errors.
    }    
\end{figure*}

\begin{figure*}[!h]
    \label{fig:exp2-results4}
    \centering
    \includegraphics[width=0.7\linewidth]{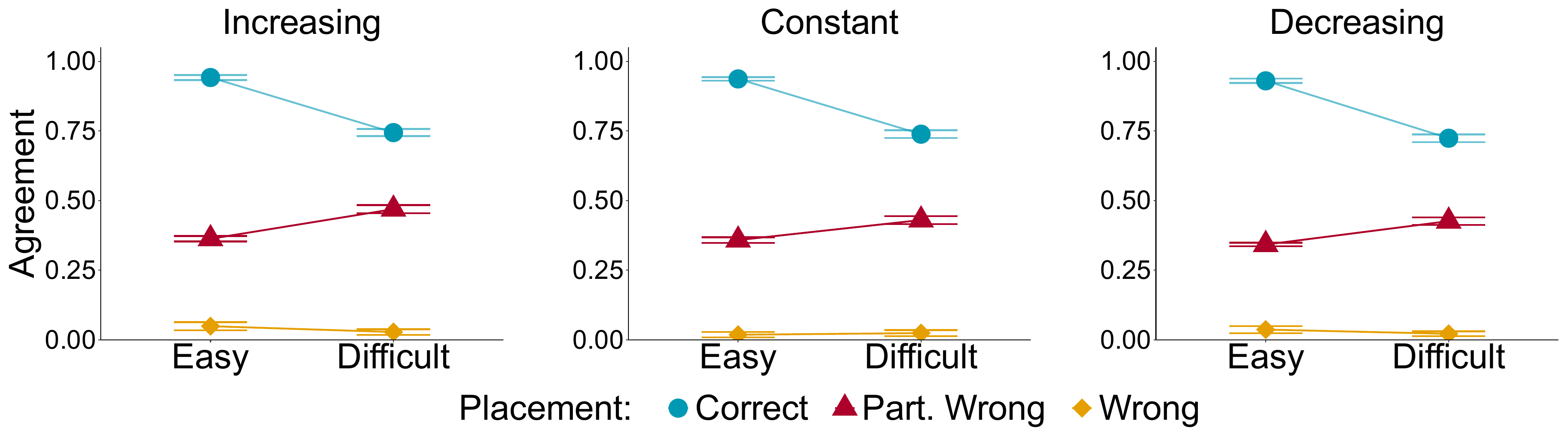}
        \caption{
        \textbf{Average agreement} in Experiment 2 divided by order condition, correctness of model’s placement (correct, partially wrong, and wrong), and image difficulty. Error bars represent standard errors.
    }
\end{figure*}

\end{document}